\documentclass[10pt,twocolumn,letterpaper]{article}

\usepackage{cvpr}      %
\usepackage{multirow}
\usepackage{multicol}
\usepackage{color, colortbl}
\usepackage{soul}
\usepackage{comment}
\usepackage{amsmath}
\usepackage{amssymb}
\usepackage{tikz}

\usepackage{pifont}%
\newcommand{\cmark}{\ding{51}}%
\newcommand{\xmark}{\ding{55}}%

\usepackage{pgfplots}
\usepackage{color}
\usetikzlibrary{spy}

\definecolor{cvprblue}{rgb}{0.21,0.49,0.74}
\definecolor{Gray}{gray}{0.9}
\definecolor{darkgreen}{RGB}{5, 120, 15}
\definecolor{lightgreen}{HTML}{9CCBB8}
\definecolor{lightred}{HTML}{E3242B}
\definecolor{lightorange}{HTML}{ED7D31}

\usepackage[pagebackref,breaklinks,colorlinks,allcolors=cvprblue]{hyperref}
\usepackage{caption}
\def\paperID{****} %
\def\confName{CVPR}
\def\confYear{2025}
\newcommand{\stwo}{{S$^2$}\xspace}
\newcommand{\method}{\textsc{HIRE}\xspace}

\title{\method: Lightweight High-Resolution Image Feature Enrichment for Multimodal LLMs
}

\author{
Nikitha SR \hspace{15pt} 
Aradhya Neeraj Mathur  \hspace{15pt}
Tarun Ram Menta\hspace{15pt}
Rishabh Jain \hspace{15pt}
Mausoom Sarkar\\
Media and Data Science Research Lab, Adobe \\
}

\begin{document}

\maketitle

\setlength{\textfloatsep}{1pt}
\begin{abstract}
The integration of high-resolution image features in modern multimodal large language models has demonstrated significant improvements in fine-grained visual understanding tasks, achieving high performance across multiple benchmarks. Since these features are obtained from large image encoders like ViT, they come with a significant increase in computational costs due to multiple calls to these encoders. In this work, we first develop an intuition for feature upsampling as a natural extension of high-resolution feature generation. Through extensive experiments and ablations, we demonstrate how a shallow feature enricher can achieve competitive results with tremendous reductions in training and inference time as well as computational cost, with upto $~1.5 \times$ saving in FLOPs.

\end{abstract}
    
\section{Introduction}
\label{sec:intro}

Multimodal LLMs (MLLMs) have emerged as a powerful paradigm, enabling LLMs to perceive, reason, and even output visual information. Seminal works such as LLaVA~\cite{liu2024improvedbaselinesvisualinstruction}, and Flamingo~\cite{alayrac2022flamingo} introduced simple and highly effective methods of introducing visual information into LLMs. 
These models leverage visual tokens obtained from a powerful vision encoder that has undergone large-scale pre-training, which are then projected into the LLM language embedding space through a projector. The popular choices of visual encoders are CLIP~\cite{radford2021learning} and its variants~\cite{zhai2023sigmoid}, and SSL based models like DINO~\cite{oquab2024dinov2learningrobustvisual}. These encoders are commonly implemented using Vision Transformers~\cite{dosovitskiy2021an}(ViTs) as the backbone architecture. \\

Recent works~\cite{liu2024improvedbaselinesvisualinstruction,chen2024dragonfly,wang2024cogvlmvisualexpertpretrained,shi2024needlargervisionmodels} have shown the immense benefits of infusing high-resolution visual information into MLLMs, enabling the perception of more fine-grained details and significantly improving performance across different tasks. 
Since the vision encoders are not pre-trained on very large image resolutions, most methods rely on splitting the input image into multiple crops, which are independently processed by the corresponding encoder and later merged~\cite{shi2024needlargervisionmodels}. Although this approach has yielded positive results, the cost of multiple ViT forward passes slows down the inference time and increases FLOPs. The computational cost becomes especially high for handling resolutions up to 2000 pixels, with methods like ~\cite{thapa2024dragonflymultiresolutionzoominencoding} requiring as many as 40 forward passes through the vision encoder. 

In order to infuse high-resolution information into the low-resolution visual representations produced by pre-trained ViTs, we explore the use of feature enrichment in the MLLM setting. 
Multiple works~\cite{suri2025lift,fu2024featup} have proposed simple methods to perform image guided deep feature upsampling, by training on simple reconstruction losses. We take inspiration from such feature upsampling works in our approach to bring in high resolution image information.

\begin{figure}
    \scalebox{0.5}{
    \includegraphics[width=1.0\textwidth]{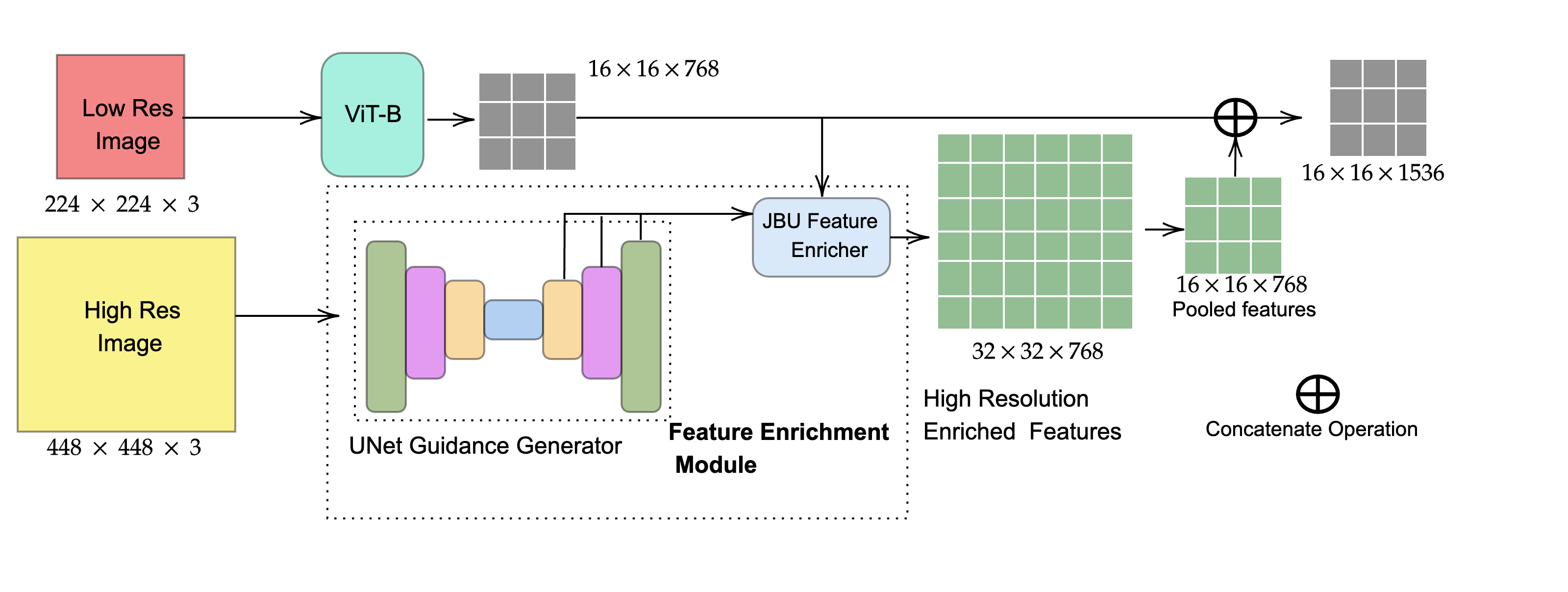}}
    \caption{The proposed method, whereby we introduce a very shallow UNet along with JBU feature enricher to generate high-resolution enriched features. The generated enriched features are then downsampled and concatenated with the lower-resolution clip features to be used further by the MLLM.
    }
    \label{fig:arch_diag}
\end{figure}

To alleviate the increased inference time and costs associated with multiple ViT forward passes for the same input image, we propose~\method, a lightweight feature enrichment pipeline for ViT features, greatly improving the efficiency of high-resolution MLLMs, while achieving competitive results with methods that use independent ViT forward passes.~\method does not require any independent pre-training, and can be used in the training pipeline of any MLLM with minimal changes. Our contributions can be summarized as follows:
\begin{itemize}
    \item We propose~\method, a novel, lightweight feature enrichment pipeline that allows high resolution information to be infused into ViT features, replacing the need for multiple ViT foward passes.
    \item~\method achieves a $35$\% reduction in FLOPs in high-resolution MLLM pipelines, while remaining competitive in performance across multiple VQA benchmarks.
    \item We demonstrate the ability of~\method to scale to larger ViT models in a more efficient manner than other high-resolution approaches for MLLMs.
\end{itemize}

\section{Method}

\subsection{Motivation and Problem Statement}

\textit{In this work, we aim to enhance the computational efficiency of multimodal large language models by reducing the need for multiple inferences with the vision encoder, a common approach in methods that incorporate high-resolution image features \cite{shi2024needlargervisionmodels,thapa2024dragonflymultiresolutionzoominencoding}. These methods typically rely on tiling high-resolution images and performing multiple forward passes through the vision encoder, leading to substantial computational overhead. As an alternative, we propose an extremely lightweight feature enricher that directly generates high-resolution visual features, thereby improving VLM performance while maintaining efficiency.}

In particular, given a high-resolution image $\mathbf{I} \in \mathbb{R}^{H \times W \times 3}$, and an image encoder $\mathcal{E}$ which processes a downsampled version of $\mathbf{I}$ ($\mathbf{i}$) to produce a feature map $\mathcal{E}(\mathbf{i}) \in \mathbb{R}^{H_f \times W_f \times D}$,~\method yields an enriched feature map $\mathcal{E}_\text{enr}(\mathbf{I})$ of resolution upto the original image size, \textit{i.e}
\begin{equation}
    \mathcal{E}_\text{enr}(\mathbf{I}) := \method(\mathbf{I}, \mathcal{E}(\mathbf{i})) \in \mathbb{R}^{H \times W \times D}
\end{equation}

\begin{figure}[!htp] 
    \centering
    \includegraphics[width=1\linewidth]{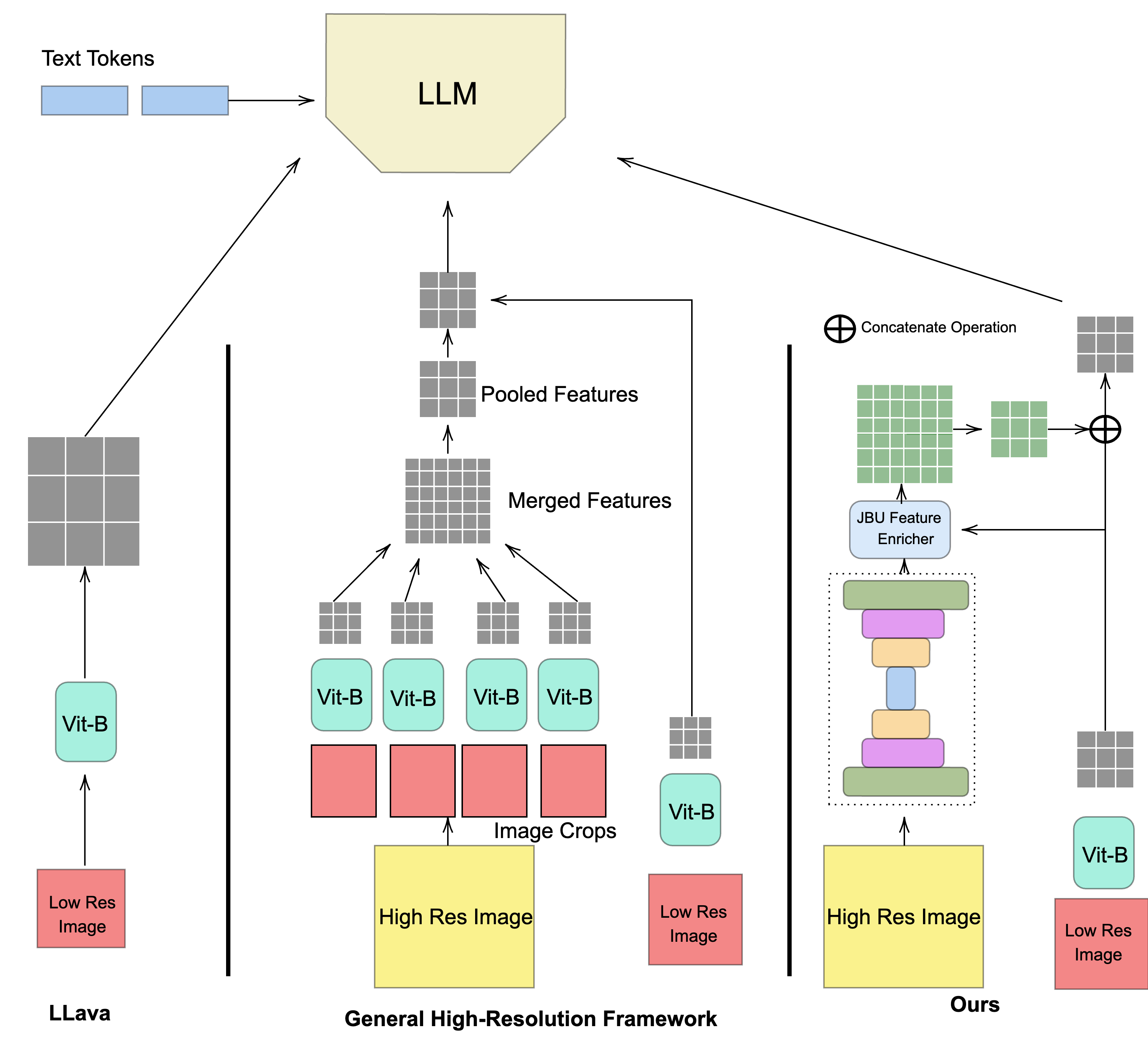}
    \caption{ We show the original LLava-based methods (left), modern high-resolution MLLMs (middle) utilize multiple ViT calls on the image crops and combine the thus obtained features, and ours (right) that uses a single shallow feature enricher to generate the high-resolution image features that are then used further.}
    \label{fig:enter-label}
\end{figure}
The proposed architecture functions as an extremely light-weight enricher, seamlessly integrating high-frequency details from the input image into the enrichment process while remaining independent of the architecture of $\mathcal{E}$.

\subsection{Approach}
Under a \textit{vision encoder, projector and LLM} setup 
 like LLaVA, provided a high-resolution image $\mathbf{I}$, we propose a lightweight enricher $\mathcal{E}_{enr}$ applied on the vision encoder that processes the low-resolution image ($\mathbf{i}$) like earlier \ref{fig:arch_diag}. 

The enricher $\mathcal{E}_{enr}$ leverages a small UNet ($U$) that acts as a semantic detail retriever from $\mathbf{I}$ along with a parameterized stack of JBU~\cite{fu2024featup} module $J$ that fuses the multi-scale features from the UNet into the $\mathcal{E}(\mathbf{i})$ and simultaneously upsamples the resolution in steps.

We use a standard UNet architecture with an encoder of $5$ downsampling modules to produce features at scales $1\times$, $1/2\times$, $1/4\times$, $1/8\times$, and $1/16\times$. These are followed by a decoder composed of a corresponding set of upsampling layers with skip connections that bring back the features to the original image resolution $672 \times 672$. The decoder stack cumulates the semantic features of the image at various scales which will act as guidance to the JBU modules ($J$) to enrich the $\mathcal{E}(\mathbf{i})$ features. $J$ consists of 5 enriching layers, each taking low-resolution image features of resolution $x$ starting from $\mathcal{E}(\mathbf{I}) \in \mathbb{R}^{24\times24\times1024}$ and guidance features of resolution $2\times x$. This guidance signal is used to generate a locality-aware joint bilateral kernel that is applied to produce the enriched features of resolution $2\times x$.

In summary, enriched features obtained through the above framework can be formulated as - 
\begin{equation}
    x_{en} = J (U(\mathbf{I}; \phi), \mathcal{E}(\mathbf{i}); \theta)
\end{equation}
where $J$ can be defined as,
\begin{equation}
    J = (\text{JBU}(\cdot, \mathbf{I}) \circ  \text{JBU}(\cdot, \mathbf{I}) \circ  ... )(U(\mathbf{I}; \phi), \mathcal{E}(\mathbf{i}))
\end{equation}
where $\circ$ is function composition. The final enriched features ($x_{en}$) are then average-pooled to the original feature scale $\mathbb{R}^{24\times24\times1024}$ and concatenated along the hidden dimension with $\mathcal{E}(i)$ before being passed to the multimodal projector. 
\begin{equation}
    \mathcal{E}_\text{enr}(\mathbf{I}) = Concat(\mathcal{E}(\mathbf{i}) , pool(x_{en}))    
\end{equation}

 This operation saves a huge computational overhead that could be caused by the quadratic scaling in the context length of the LLM if the high-resolution enriched features were used directly by the projector for tokenization. 
\\
The UNet module helps to bring stronger signals from $\mathbf{I}$ into low-resolution features, with the total size of the enricher module still being only \textbf{~250k} parameters as shown in Fig.~\ref{fig:arch_diag}. The effect of incorporating \textit{U} is discussed in Table \ref{tab:ablations}.

\begin{figure*}[!htp]
    \centering
    \scalebox{0.65}{\includegraphics[width=0.9\linewidth]{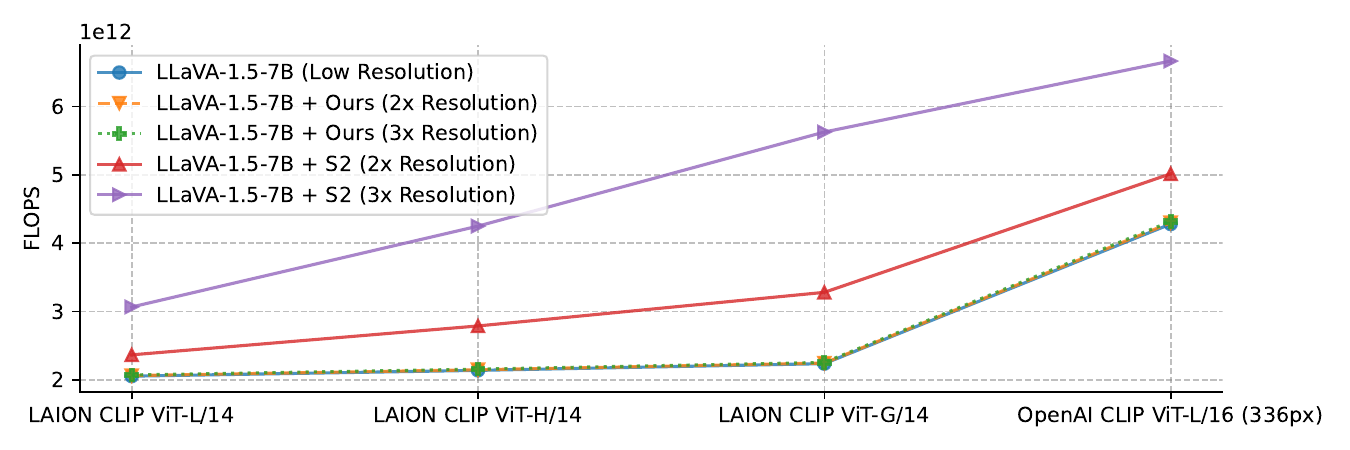}}
    \caption{Required FLOPs for first token generation for LLaVA1.5, LLaVA1.5-\stwo, and LLaVA1.5-\method combined with ViTs of varying size and resolution.~\method adds negligible overhead over vanilla LLaVA1.5, and is much more efficient than \stwo.}
    \label{fig:scaling}
\end{figure*}

\section{Experiments and Analysis}
\label{sec:experiments}
In order the evaluate the efficacy of~\method, we first begin by integrating~\method into the training pipeline of LLaVA-1.5. We evaluate~\method~on a variety of visual detail understanding and question answering tasks.
We then examine the compute requirements of~\method when scaling to larger ViTs (Sec.~\ref{sec:scaling}) showcasing the efficacy of ~\method over \stwo.

\begin{table}[!htp]
    \centering
    \scalebox{0.65}{
    \begin{tabular}{cc|c}
     \toprule   
     Model & Resolution & TFLOPs $\downarrow$ \\
     \midrule
     LLaVA1.5-7B & $336$ & $4.278$\\
     \midrule
     LLaVA1.5-7B-\stwo & $672$ & $5.013$\\
    LLaVA1.5-7B-\method (ours) & $672$ & $\mathbf{4.296}$\\
     \midrule
     LLaVA1.5-7B-\stwo & $1008$ & $6.664$\\
    LLaVA1.5-7B-\method (ours) & $1008$ & $\mathbf{4.317}$\\
    \bottomrule
    \end{tabular}}
    \caption{Comparison of FLOPs utilization for first token generation in each model. Prompt consists of a single image with a single text token in all cases.~\method consistently adds much lower computational overhead as compared to the multiple ViT calls in \stwo, while maintaining strong performance as demonstrated in Table.~\ref{tab:vllm}.}
    \label{tab:flops}
\end{table}

\begin{table}[!htp]
\scriptsize
    \scalebox{0.75}{
    \begin{tabular}{|c|c|c|c|c|c|}
        \hline
        Model &  Pixel  & Image Dense  & Animal  Detection & Hallucination & Overall  \\
        & Localization & Captioning & Keypoint & &\\
         \hline
         LLaVA1.5-7B-S2 & 20 & 31.58 &  30 & 56.25 &49 \\
         LLaVA1.5-7B-\method & \textbf{30}  & \textbf{42.11} & \textbf{45} &  \textbf{61.25} & \textbf{49.9}\\
         \hline
    \end{tabular}}
    \caption{ Results on MMT-Bench show clear improvements across multiple subtasks on fine grained understanding, with both individual and overall scores presented above.}
    \label{tab:mmt_f}
\end{table}

\begin{table}[!htp]
    \centering
    \begin{scriptsize}
    \scalebox{0.85}{
    \begin{tabular}{lp{0.03\textwidth}p{0.04\textwidth}|p{0.03\textwidth}p{0.03\textwidth}p{0.03\textwidth}p{0.03\textwidth}p{0.03\textwidth}|p{0.04\textwidth}p{0.03\textwidth}p{0.03\textwidth}p{0.03\textwidth}p{0.05\textwidth} }
        \toprule
        \multicolumn{3}{c}{}&&\\
        \multirow{2}{*}{Model} & \multirow{2}{*}{Res.} & \multirow{2}{*}{\#Token}  & DocVQA & InfoQA & SciQA & SEED-Img  \\
        
        \midrule
       
        LLaVA1.5-7B & 336 & 576 & 18.6 & 19.1 & 66.8 & 66.1 \\

        LLaVA1.5-7B-\stwo~ & 1008 & 576  &\textbf{26}&\textbf{21.9}&69.5&67.9\\

        \rowcolor{lightgreen!40} \textbf{LLaVA1.5-7B-\method} &  672 & 576 &20.7&18.93&\textbf{70.1}&67.65\\
        \hline
        LLaVA1.5-7B-Qwen & 1008 & 576 &22.07&20.98&75.86&68.93\\

        \rowcolor{lightgreen!40} \textbf{LLaVA1.5-7B-Qwen-\method} &  672 & 576 &22.17&19.81&75.31&\textbf{69.75}\\
        
        \midrule
        \midrule

        &&& \multirow{2}{*}{VQA$^{\text{T}}$} &
        \multirow{2}{*}{
        Viz}  & \multirow{2}{*}{GQA} & \multirow{2}{*}{MMVet}\\
        & & & & & \\
        \midrule
        LLaVA1.5-7B~\cite{liu2024improvedbaselinesvisualinstruction} & 336 & 576 &  58.2 & 50.0  & 62  & 30.5 \\

        LLaVA1.5-7B-\stwo~\cite{shi2024needlargervisionmodels} & 1008 & 576 & \textbf{61} &  50.1 & 63 & 31.6\\

        \rowcolor{lightgreen!40} \textbf{LLaVA1.5-7B-\method} &  672 & 576 &  57.8 & \textbf{51.0} & \textbf{63.1} & 32.9 \\
        \hline
        LLaVA1.5-7B-Qwen & 1008 & 576 & 57&46.25&61.89&35\\

        \rowcolor{lightgreen!40} \textbf{LLaVA1.5-7B-Qwen-\method} &  672 & 576 &57.58&\textbf{46.8}&61.74&\textbf{35.2}\\

        \midrule
        \midrule

        &&&\multirow{2}{*}{V$^\ast_{\text{Att}}$} & \multirow{2}{*}{V$^\ast_{\text{Spa}}$} & \multirow{2}{*}{VQA$^{\text{v2}}$ } & Pope-Adv\\
        & & & & \\
        \midrule
        LLaVA1.5-7B~\cite{liu2024improvedbaselinesvisualinstruction} & 336 & 576 & 43.5 & 56.6 & 78.5 & 84.2 \\

        LLaVA1.5-7B-\stwo~\cite{shi2024needlargervisionmodels} & 1008 & 576 & 46.1\textsuperscript{T} & 61.8 & 80.0 &85.6\\

        \rowcolor{lightgreen!40} \textbf{LLaVA1.5-7B-\method} &  672 & 576 & \textbf{46.1} & 60.5 & 79.31 &85 \\

        \bottomrule
        \bottomrule

    \end{tabular}}
    \end{scriptsize}
    
    \caption{\scriptsize \textbf{Results on MLLM.} We evaluate three types of benchmarks: Visual detail understanding (V$^*$), visual question answering (DocVQA, InfoQA, SciQA, SEED, Pope) and MLLM benchmarks (MM-Vet, VizViz, MMT-Bench).
    \textcolor{gray}{\textsuperscript{T} All scores were reproduced using the provided checkpoint and evaluation scripts from the authors.  
    }}
    \label{tab:vllm}
    \vspace{0.5em}
\end{table}

LLaVA-1.5-\method is trained in a 2-stage pre-training and finetuning framework with CLIP-336 as the vision encoder and a 2-layer MLP as the projector, exactly following the same hyperparameters as LLaVA-1.5. The LLM (Vicuna-7B/Qwen-7B) is trained with LoRA~\cite{hu2021lora} to reduce the memory requirement. During the pretraining stage, we unfreeze both the projector and the~\method module. During the finetuning stage, we keep the same modules unfrozen and subsequently add trainable LoRA modules to the LLM layers. The results of various benchmark evaluations are presented in Table~\ref{tab:vllm}.

\setlength{\belowcaptionskip}{1pt} 
\begin{table}[!hpt]
    \centering
    \setlength{\tabcolsep}{3pt}
    \scalebox{0.65}{
    \begin{tabular}{lcc|ccc}
        \toprule
         Experiment & U-Net & \textit{lr} & V$^\ast_{\text{Att}}$ & V$^\ast_{\text{Spa}}$ & V$^\ast_{\text{Avg}}$ \\
         \midrule
         Bi-Cubic Interpolation & - & - & 45.2 & 53.9 & 48.7 \\
         \midrule
         LLaVA1.5-7B-\method & \cmark & 2e-5 & 46.1 & \textbf{60.1} & \textbf{51.8} \\
          & \xmark & 2e-5 & 44.3 & 57.9 & 49.7\\
          & \cmark & 5e-3 & \textbf{46.8} & 55.3 & 50.3\\
           & \xmark & 5e-3 & 41.73 & 55.26 & 47.12 \\
         
         \bottomrule
    \end{tabular}}
    \caption{Effect of key components of~\method on V$^\ast$. V$^\ast$ is a challenging benchmark that requires very performant high-resolution image features for answering via an MLLM. Higher is better. Best result in \textbf{bold}.}
    \label{tab:ablations}
\end{table}

Table~\ref{tab:flops} demonstrates the low inference cost of~\method using Fvcore library for FLOP analysis. We used 32 A100s for training and one A100 for inference. Benchmarks such as V$^\ast$, VQA$^T$, and GQA are of particular interest, since they require high-resolution information to answer, though all benchmarks are aided by additional high-resolution information.  LLaVA1.5-7B-\method outperforms the vanilla variant across multiple benchmarks like the high-resolution dependent visual detail understanding tasks, V$^\ast_{spa}$ ($6$\% relative $\uparrow$) and V$^\ast_{Att}$ ($7$\% relative $\uparrow$), as well as other VQA benchmarks like VizWiz, ScienceQA, VQA$^{v2}$, SEED, and GQA despite using a lower resolution ($672$ vs.$1008$), adding minimal parameters ($0.25$M) and significantly small computational overhead ($4.278$ vs $6.664$ TFLOPs). We additionally show results on fine-grained MMT-Bench tasks in Table \ref{tab:mmt_f} using Vicuna 7B where our method outshines \stwo by $\sim10\%$ on pixel-level localization and dense captioning,  and $\sim15\%$ for animal keypoint detection, which are fine-grained tasks requiring good high-resolution features. An improvement of $\sim5\%$ is observed in the hallucination task, indicating a reduction in hallucination. This demonstrates the success of~\method in side-stepping the need for multiple ViT calls to obtain high-resolution image features.  We also compare the total inference time on the VQA$^{\text{T}}$ test set. \stwo takes $43$ mins which is ($>2\times$) the inference time compared to LLaVA-1.5 while HIRE takes $23.2$ mins which is an increase of only ($0.1\%\uparrow$) on the entire test set compared to LLaVA1.5.

Areas where~\method falls behind are VQA$^T$ and InfographicsVQA where our variant performs below both the \stwo and vanilla variants. We posit that this is due to a lack of quality in enriching text-heavy images and leave it for future research. We also perform ablations on using UNet in the enricher and finetuning with different learning rates in Table \ref{tab:ablations}. It can be seen that using the UNet brings in strong semantic features to enrich the vision encodings leading the better downstream performance in V$^*$ benchmark.

\subsection{Scaling with \method}
\label{sec:scaling}

The visual encoder plays a key role in the overall downstream performance of MLLMs. 
In this section, we study the scalability of~\method to enrich features from larger ViT backbones, which is an important capability to build larger, powerful MLLMs. We present the results in Fig.~\ref{fig:scaling}. Across ViTs of different scales (Large, Huge, Giant) and resolutions (224, 336),~\method adds negligible overhead compared to the vanilla variant, whereas \stwo incurs significantly higher FLOP usage due to multiple ViT forward passes for the first token generation. While the compute overhead of~\method remains constant across all ViTs, the compute overhead for \stwo increases as the size/resolution of the ViT is increased.

\section{Conclusion}
In this work, we present a generic feature enrichment technique that can be applied to Multimodal LLMs that depend on multi-resolution images for achieving higher performance. Our method not only saves tremendously on the compute during both training and inference but also surpasses the performance of the state-of-the-art method \stwo on various multimodal tasks while demonstrating the capability to extend seamlessly for larger ViTs.

{
    \small
    \bibliographystyle{ieeenat_fullname}
    \bibliography{main}
}

\end{document}